\documentclass[fleqn,10pt]{wlscirep}

\usepackage[utf8]{inputenc}
\usepackage[T1]{fontenc}
\usepackage{multirow}
\usepackage{subcaption}
\usepackage{graphicx}
\usepackage{url}
\usepackage{amsmath}
\usepackage{mathtools}

\title{Classifying patient voice in social media data using neural networks: A comparison of AI models on different data sources and therapeutic domains.}

\author[1,+]{Giorgos Lysandrou}
\author[1,+]{Roma English Owen}
\author[1]{Vanja Popovic}
\author[1]{Grant Le Brun}
\author[1]{Beatrice Alex}
\author[1,*]{Elizabeth A. L. Fairley}
\affil[1]{Talking Medicines Limited, SC447227, 25 Blythswood Square, G2 4BL, Glasgow, Scotland, UK}
\affil[*]{elizabeth@talkingmedicines.com}

\affil[+]{these authors contributed equally to this work}

\keywords{patient voice, cardiovascular, oncology, immunology, neurology, social media, classification, neural network, transformer, language model}

\begin{abstract}

It is essential that healthcare professionals and members of the healthcare community can access and easily understand patient experiences in the real world, so that care standards can be improved and driven towards personalised drug treatment. Social media platforms and message boards are deemed suitable sources of patient experience information, as patients have been observed to discuss and exchange knowledge, look for and provide support online. This paper tests the hypothesis that not all online patient experience information can be treated and collected in the same way, as a result of the inherent differences in the way individuals talk about their journeys, in different therapeutic domains and or data sources.

We used linguistic analysis to understand and identify similarities between datasets, across patient language, between data sources (Reddit, SocialGist) and therapeutic domains (cardiovascular, oncology, immunology, neurology). We detected common vocabulary used by patients in the same therapeutic domain across data sources, except for immunology patients, who use unique vocabulary between the two data sources, and compared to all other datasets. We combined linguistically similar datasets to train classifiers (CNN, transformer) to accurately identify patient experience posts from social media, a task we refer to as patient voice classification. The cardiovascular and neurology transformer classifiers perform the best in their respective comparisons for the Reddit data source, achieving F1-scores of 0.865 and 1.0 respectively. The overall best performing classifier is the transformer classifier trained on all data collected for this experiment, achieving F1-scores ranging between 0.863 and 0.995 across all therapeutic domain and data source specific test datasets.

Our findings show that linguistic and statistical text similarity analysis is essential in identifying and understanding commonalities in language between patient groups. These results allude to even starker differences in the way patients speak at a disease level and across additional therapeutic domains. The analysis outcome can be leveraged to aid dataset creation, as well as used to interpret classifier evaluation scores. With few exceptions, the pre-trained transformer text classifier, trained on more data by combining datasets, was overall the best performing classifier in these experiments. This confirms our initial hypothesis, demonstrating the importance of understanding and identifying common patient language for optimal patient voice classification.

\end{abstract}
\begin{document}

\flushbottom
\maketitle
\thispagestyle{empty}
% \pagestyle{empty}

% \noindent Please note: Abbreviations should be introduced at the first mention in the main text – no abbreviations lists. Suggested structure of main text (not enforced) is provided below.

\section*{Introduction}

% The Introduction section, of referenced text\cite{Figueredo:2009dg} expands on the background of the work (some overlap with the Abstract is acceptable). The introduction should not include subheadings.

% \subsection*{Introduction and Motivation}
\subsection*{Background and Motivation}

There is a critical need for healthcare systems all round the world to provide better outcomes for patients. A significant proportion of our healthcare systems' deficiency to provide adequate care can be attributed to the generalized nature of services and drugs that are provided. If patients received more personalized care, i.e. improved access to the right drugs, at the right time, the right dose and formulation, we could expect outcomes to improve vastly \cite{grissinger2010five}. By shifting our focus from primarily listening to health care professionals and members of the medical community, to listening more to patients, we can better understand how patients are taking and experiencing their drugs on a day-to-day basis, leading to improved patient outcomes through more personalized care and the development of more appropriate treatments for all.

However, capturing patient experiences is not as simple as it may seem, and conventional methods for doing so may not provide the full picture and yield inadequate or inaccurate results. For instance, extensive patient experience information is obtained through interactions with healthcare professionals (HCPs). Though this is a valuable source of information, often the environment patients are in throughout these discussions can result in less descriptive conversations, with the patients reporting that the HCPs fail to grasp their perspective, resulting in a poor care experience \cite{leonard2017exploring}. Another well known source of patient information is patient focus groups, with the patients' interactions being recorded to generate data from their perspective on healthcare related topics, like their experience using certain drugs. These groups are self-selecting and by the nature of this could result in biased feedback coming only from individuals within a particular socioeconomic segment - one that is not representative of the entire group of patients taking a given drug.

An alternative medium for collecting information on patients' experiences is social media. Patients are observed to talk about their health, sharing sensitive personal information, enabled by the distance and anonymity online \cite{colineau2010talking}\cite{antheunis2013patients}. Patients on social media are not hindered by the presence of HCPs, and truly express their experiences descriptively in their own language, compared to when patients are within a clinical and health care environment. This can affect to an extent the information collected by HCPs contained in clinical and electronic health datasets, which while rich in patient factual biomedical information, does not contain patient behavioural and emotional information, and can often be hard to obtain. Lastly, compared to more traditional data for sourcing patient information, such as market research, pharmaceutical representatives feedback and phone call data, patients have easier access to social media, which results in patient information coming from a wider range of socioeconomic and geographic backgrounds, reducing the bias of the aforementioned data sources.

Due to this realization, over the last few years there has been an apparent increase in research on analysing social media data for patient experience information. This was likely driven not only by advancements in data capture and analysis (through Application Programming Interfaces (API’s) and improvements in Artificial Intelligence (AI) technology) but also by a significant increase in relevant social media data available, in part fuelled by the worldwide COVID-19 pandemic. Websites such as Reddit (\url{https://www.reddit.com/}), have an international user base, with users organising themselves in groups related to particular topics, like a disorder or a treatment. An example would be the subreddit r/eczema (\url{https://www.reddit.com/r/eczema/}), where a community of users share their experiences and knowledge around eczema, in the therapeutic domain of immunology.

Our hypothesis is that not all patient experience information can be treated and collected in the same way as a result of the inherent differences in the way individuals talk about their experiences depending on the therapeutic domain and data sources in which they are talking. 

The goal of this research and development project is to test this hypothesis by detecting and understanding the linguistic differences in how patients talk about their conditions and experiences, and to aid more optimal patient voice classification. In order to do this, we collected data across social media (Reddit) and message boards (SocialGist - accumulator of message board websites) and therapeutic domains (cardiovascular, oncology, immunology and neurology (COIN)). Starting from our hypothesis, initial top level analysis shows the subsets of the data to be particularly different in nature, requiring further exploration. Experiences related to cardiovascular conditions appeared to be largely numerical and referred to blood pressure readings, dosages, heart rates etc. Experiences related to oncology referred to specific genes and genomic sequences. Accounts on immunology appeared to be highly descriptive in nature with ample use of adjectives, and finally patient accounts on neurology were largely based on emotions and feelings rather than physical symptoms. It is precisely because of these differences that these therapeutic domain areas were selected for study.

Previous work in this area includes Alex et al. (2021) \cite{alex2021classifying}, who investigated the identification of patient voice in social media posts. Their work focussed on extracting posts from social media and categorizing them either as having been written by a patient, healthcare professional or other. Alex et al. presented the performance of their classifier model (a convolutional neutral network classifier trained on English language Reddit and Twitter data across two therapeutic domains: cardiovascular and skin conditions) and the importance of training two separate models – one for each data source. They report a minor increase in performance by training per therapeutic domain.

In this paper, we broaden the scope of this research by:

\begin{itemize}
    \item Identifying an appropriate linguistic and text similarity analysis to guide and explain the patient voice classification,
    \item Understanding the commonalities and differences between the ways patients express themselves, about different health conditions, depending on therapeutic domain and data source,
    \item Determining the optimal machine learning (ML) classification methodology to classify posts belonging to patients, amongst different therapeutic domains and data sources.
\end{itemize}

% ////////////////////////////////////////////////////////////////////////////////
\subsection*{Related Work}

The digitization of healthcare, technological advancements in AI and the influx of patient voice online because of global events such as the COVID-19 pandemic, have all been catalysts in the use of AI in healthcare. Jiang et al. (2017) in their survey emphasized the benefits of applying AI on structured and unstructured healthcare data. They motivate the use of Natural Language Processing (NLP) and Neural Network (NN) based learning for unstructured data, such as clinical notes and medical journals, which are the tools used in the experiments conducted in this paper. Jiang et al. also referenced the main therapeutic domains that use AI technology at present (oncology, neurology and cardiology) and review the utility of AI applications in these areas for early detection and diagnosis, treatment and outcome prediction of patients that are at risk of strokes. They also review the difficulties of using AI in practice \cite{jiang2017artificial}.

% . . . . . . . . . . . . . . . . . . . . . . . . . . . . . . . . . . . . . . . . 
\subsubsection*{Natural Language Processing in Healthcare}

Hudaa et al. (2019), emphasized the use of NLP techniques in clinical informatics research. In particular, they report on the importance of using NLP to structure unstructured information, through document classification and feature extraction. A vast amount of healthcare information assets are not structured, or follow incompatible frameworks, but are all written in natural language, which can be understood with the recent advancements in Large Language Models (LLMs). Hudaa et al. believe that patient-healthcare provider relationships could be improved, as a result of NLP providing a lift on the analysis of healthcare information, making the research of healthcare professionals easier \cite{hudaa2019natural}.

Lavanya et al. (2021) also touched on the unstructured nature of social media posts and emphasized that optimisation needs to be done to find the best solution for retrieving this information. This is especially important in light of how quickly social media has become the go-to place for discussing and sharing opinions about all manner of things, including healthcare experiences and drugs.  Their work explored the benefit of using deep learning techniques applied to NLP text classification of data collected from healthcare related social media networks \cite{lavanya2021deep}.

% . . . . . . . . . . . . . . . . . . . . . . . . . . . . . . . . . . . . . . . . 
\subsubsection*{Patient Voice Detection}

Pattisapu et al. (2017) looked for patterns in social media posts, specifically blog posts and tweets, to infer the medical persona of the author. Through their meticulous feature engineering, which was shown to outperform manual feature selection, they extracted useful information from pre-trained word embeddings in patients' posts. They argued that this type of analysis is of great importance for the marketing of drugs and for monitoring drug safety \cite{pattisapu2017medical}.

Dai et al. (2017) examined clustering social media posts embeddings, as an alternative to bag-of-words classification, as they believed the latter to be inferior at comprehending the intended meaning of social media posts. Interestingly, the clustering of word embeddings method they reported on is entirely unsupervised, meaning that no labelled training data was used in this particular study. They reported a classification accuracy of 87.1\% using this method \cite{dai2017social}.

Alex et al. (2021) provide an extensive related work section on patient voice detection in their work "Classifying patient and professional voice in social
media health posts." \cite{alex2021classifying}, we are merely expanding on that.

% . . . . . . . . . . . . . . . . . . . . . . . . . . . . . . . . . . . . . . . . 
\subsubsection*{Linguistic Analysis of Patient Language}

Lu et al. (2013), in an attempt to better understand patients, performed topic analysis in online patient communities \cite{lu2013health}. Their findings showed that the emerging topics included discussions on symptoms, examinations, drugs, procedures and complications, with a significant difference among the topics captured between different communities for different diseases.

While Dreisbach et al. (2019) provided a review of publications that reported on NLP used in relation to symptom extraction and electronic patient-authored text. Their report looked at studies that have been used to detect clinical symptom categories, from Twitter and online community forums, primarily in the areas of diabetes, cancer and depression \cite{dreisbach2019systematic}.

% ////////////////////////////////////////////////////////////////////////////////
\subsection*{Data}

For the purpose of the experiments conducted in this paper, we automatically collected social media posts from Reddit (\url{https://www.reddit.com/}) and message board posts from SocialGist (\url{https://socialgist.com/}), reporting on cardiovascular, oncology, immunology and neurology conditions. While Reddit is a single data source, split into user-created communities, or subreddits, SocialGist is an accumulator of data sources, collecting posts from websites of various communities.
We provide the list of subreddits and SocialGist websites in the file Supplementary Material 1.

% . . . . . . . . . . . . . . . . . . . . . . . . . . . . . . . . . . . . . . . . 
\subsubsection*{Data Collection and Preparation}

Reddit posts were gathered using the Pushshift Reddit API (\url{https://github.com/pushshift/api}, this was used in order to obtain historical posts too as well as those that were up-to-date). We collected Reddit posts by searching a curated selection of subreddits, for a set of specific search terms (drugs) across four therapeutic domains - cardiovascular, oncology, immunology and neurology. The same set of search terms was used to gather posts from message boards using the SocialGist API. It should be noted that we did not formally assess the relevance of posts to each domain, however previous work indicates that a manually curated list of search terms generally leads to high recall and precision \cite{llewellyn2015extracting}.

Duplicate posts were removed from the dataset (where a duplicate is regarded as a post with matching text body or matching identifier to one already gathered). In total, 14,693 posts were collected, 7,211 from Reddit and 7,482 from the social media aggregator SocialGist (broken down in Table~\ref{tab:experiment_data_volumes} by data source and domain).

% . . . . . . . . . . . . . . . . . . . . . . . . . . . . . . . . . . . . . . . . 
\subsubsection*{Manual Annotation}

The collected posts were manually annotated by a group of annotators. Using the open-source annotation tool, Doccano (\url{https://doccano.github.io/doccano/}), the trained annotators, following comprehensive annotation guidelines (developed and refined over a period of time, and applicable for this experiment), labelled each post by post type. The experiments reported in this paper, focus on document-level annotation and classification. Posts describing the first-hand experience of a patient were annotated as patient voice, while all other posts, such as Health Care Professionals, News Articles, amongst other types, were annotated as Not Relevant. Some synthetic examples of each label, are as follows:

\begin{itemize}
    \item \textsc{Patient Voice}: "I'm taking MTX and imraldi at the moment, so far so good." 
    % \item \textsc{Professional Voice}: "One of my old patients used to take 10mg eliquis instead of 5mg. His heart rate was..." 
    \item \textsc{Not Relevant}: "MHRA due to approve new RA drug.", "One of my old patients used to take 10mg eliquis instead of 5mg. His heart rate was..." 
\end{itemize}

The data was split using and 80\% training and 20\% validation data ratio. 80\% of the data was used to train the classifier models, while the remaining 20\% of the data was used to evaluate the best performing trained classifiers. Additional data, equal to the 20\% validation data, was collected as a test dataset, to evaluate the classifiers' performance and report their precision, recall and F1-score metrics. No validation or test data was part of the classifiers' train data. To avoid a biased split, the data is initially shuffled, with a reproducible random seed, and the label distribution between the train and test datasets is the same. Table~\ref{tab:experiment_data_volumes} details the distribution of label counts across all domains and both data sources of the data.

\begin{table}[h]
\centering
\begin{tabular}{llllllll}
\hline
\multirow{2}{*}{Therapeutic Domain} & \multirow{2}{*}{Data Source} & \multicolumn{3}{c}{Patient Voice} & \multicolumn{3}{c}{Not Relevant} \\
                                &            & Train & Validation & Test & Train & Validation & Test      \\
\hline
\multirow{2}{*}{Cardiovascular} & Reddit     & 884   & 216 & 186  & 319   & 109 & 112  \\
                                & SocialGist & 876   & 224 & 236  & 405   & 95  & 61   \\
\multirow{2}{*}{Oncology}       & Reddit     & 887   & 213 & 276  & 390   & 106 & 24   \\
                                & SocialGist & 876   & 224 & 275  & 401   & 99  & 23   \\
\multirow{2}{*}{Immunology}     & Reddit     & 881   & 219 & 277  & 399   & 101 & 22   \\
                                & SocialGist & 872   & 228 & 261  & 407   & 93  & 29   \\
\multirow{2}{*}{Neurology}      & Reddit     & 882   & 218 & 187  & 233   & 59  & 11   \\
                                & SocialGist & 871   & 226 & 144  & 362   & 86  & 108  \\
\hline
\end{tabular}
\caption{Experiment data volumes across data sources, therapeutic domains, train, validation and test splits, for the classes of Patient Voice and Not Relevant social media posts.}
\label{tab:experiment_data_volumes}
\end{table}

Table~\ref{tab:experiment_combined_data_volumes} details the distribution of label counts across the combined datasets. Datasets across two different data sources, in the same therapeutic domain, have been combined into a single therapeutic domain dataset, e.g. the cardiovascular dataset, contains cardiovascular posts both from Reddit and SocialGist. Datasets across multiple therapeutic domains, from the same data source, have been combined into a single data source dataset, e.g. Reddit COIN, 
% short for Cardiology, Oncology, Immunology and Neurology, 
containing all Reddit posts across all therapeutic domains. Lastly, a single All dataset, contains all posts, across all data sources and therapeutic domains. When combining datasets, the train, validation and test datasets are combined with their respective train, validation and test datasets.

\begin{table}[h]
\centering
\begin{tabular}{lllllll}
\hline
\multirow{2}{*}{Combined Datasets} & \multicolumn{3}{c}{Patient Voice} & \multicolumn{3}{c}{Not Relevant} \\
                & Train & Validation  & Test & Train & Validation & Test \\
\hline
Cardiovascular  & 1760  & 440  & 422  & 724   & 204 & 173  \\
Oncology        & 1763  & 437  & 551  & 791   & 205 & 47   \\
Immunology      & 1753  & 447  & 538  & 806   & 194 & 51   \\
Neurology       & 1753  & 544  & 331  & 595   & 145 & 119  \\
Reddit COIN     & 3534  & 866  & 926  & 1341  & 375 & 169  \\
SocialGist COIN & 3495  & 902  & 916  & 1575  & 373 & 221  \\
All             & 7029  & 1768 & 1842 & 2916  & 748 & 390  \\
\hline
\end{tabular}
\caption{Experiment combined data volumes across datasets with combined data sources, therapeutic domains, train, validation and test splits, for the classes of Patient Voice and Not Relevant social media posts.}
\label{tab:experiment_combined_data_volumes}
\end{table}

% ≡≡≡≡≡≡≡≡≡≡≡≡≡≡≡≡≡≡≡≡≡≡≡≡≡≡≡≡≡≡≡≡≡≡≡≡≡≡≡≡≡≡≡≡≡≡≡≡≡≡≡≡≡≡≡≡≡≡≡≡≡≡≡≡≡≡≡≡≡≡≡≡≡≡≡≡≡≡≡≡

\section*{Results}

% Up to three levels of \textbf{subheading} are permitted. Subheadings should not be numbered.

% \subsection*{Subsection}

% Example text under a subsection. Bulleted lists may be used where appropriate, e.g.

% \begin{itemize}
% \item First item
% \item Second item
% \end{itemize}

% \subsubsection*{Third-level section}
 
% Topical subheadings are allowed.

% ≡≡≡≡≡≡≡≡≡≡≡≡≡≡≡≡≡≡≡≡≡≡≡≡≡≡≡≡≡≡≡≡≡≡≡≡≡≡≡≡≡≡≡≡≡≡≡≡≡≡≡≡≡≡≡≡≡≡≡≡≡≡≡≡≡≡≡≡≡≡≡≡≡≡≡≡≡≡≡≡

% ////////////////////////////////////////////////////////////////////////////////
\subsection*{Inter-Annotator Agreement}

The Inter-Annotator Agreement (IAA) is computed for the labels assigned to each post by different annotators to understand the alignment of the annotators. This is also informative on the effectiveness of the annotation guidelines, as well as the difficulty of the classification task. The IAA scores determine the upper bound of performance a classifier model can realistically achieve, if it is modelling human classification.

IAA scores were produced for each pair of annotators, for a total of 12 annotators annotating 2,388 randomly selected posts, across all four therapeutic domains - COIN. Standard precision, recall and F1-score were calculated for each pair, for each label class, weighted and macro averaged. Cohen's Kappa score was also calculated for each pair of annotators, as a measure that takes into account the probability of the annotators agreeing or disagreeing by chance, with a resulting score that is more representative and robust than precision, recall and F1-score \cite{warrens2015five}. The average Cohen's Kappa score between annotators is 0.773, which can be interpreted as notably substantial agreement, also supported by the high precision, recall and F1-scores observed. The IAA scores can be seen in detail in Table~\ref{tab:iaa}.

\begin{table}[h]
\centering
\begin{tabular}{lll}
\hline
\multirow{2}{*}{Metric} & Weighted & Macro    \\
                        & Averages & Averages \\
\hline
Precision & 0.956             & 0.835          \\
Recall    & 0.947             & 0.935          \\
F1-Score  & 0.952             & 0.883          \\
\hline
\end{tabular}
\caption{Weighted and macro averaged mean precision, recall and F1-score for Inter-Annotator Agreement, across the dataset and all annotators.}
\label{tab:iaa}
\end{table}

% ////////////////////////////////////////////////////////////////////////////////

\subsection*{Linguistic Analysis}

Manual analysis and review of randomly selected posts from each therapeutic domain and data source specific dataset, shows differences in the language patients use to describe their experiences. Overall, SocialGist patients describe their experiences using more words (longer posts), compared to Reddit patients, while also using a richer vocabulary. Our analysis provides us with a deeper understanding of the language in each therapeutic domain, with some examples showcasing these variations, e.g.:

\vspace{0.5cm}

\begin{itemize}
    \item \textsc{Cardiovascular:} \textit{“I had a blood test for my D levels. They were scarily low the year of my heart attack (2016). The doctor put me on 1000mg a day of D3.”}:
    \item \textsc{Oncology:} \textit{“I'm in a similar situation to you! I'm 32, was diagnosed in April, I have HER2+ invasive Stage 2B grade 2 with lymph node involvement.”}:
    \item \textsc{Immunology:} \textit{“I now take Cosentyx. However, after 3 months I have noticed what looks either like psoriasis on soles of my feet or could be athlete’s foot. After 15 weeks, I now have the same peeling all over palms, between fingers, and backs of fingers.”}:
    \item \textsc{Neurology:} \textit{“I have taken quetiapine, abilify and olanzapine. Olanzapine didn’t help at all and I always feel a bit nervous and anxious on the quetiapine.”}:
\end{itemize}

\vspace{0.3cm}

% Cardiovascular related posts tend to mention dosages, tests and cardiovascular related events. Posts on oncology typically mention specific proteins and descriptive medical language. Immunology specific posts usually include descriptive language of physical symptoms, and neurology specific posts often contain descriptive language of emotions and feelings.

Cardiovascular related posts tend to mention dosages, tests and cardiovascular related events for instance, heart attacks and strokes. Oncology posts typically contain highly specific information about distinct proteins and genes relating to these conditions as well as stage of disease e.g. HER2+, ER-, Stage IV. Immunology patient posts usually include descriptive language of physical and mental symptoms e.g. dry, itchy, crawling. Neurology posts, though still descriptive in nature hone in specifically on emotions and feeling experienced by the patient both when receiving treatment and when not.

To accompany this manual analysis on the experiment's data, an appropriate statistical text similarity analysis is necessary to bridge the manual analysis and the classifier models. The choice of the statistical text similarity analysis is determined by the classification methodology of the two classifiers. The two classifiers in this experiment, while different in architecture, both classify text using the same bag-of-words approach.

This means that each word in a text is represented by a vector created by the language models, and each text is represented by the combination of the word representation vectors of the words that make up the text. Each word is embedded and encoded into a context-depended vector, with the context derived from the meaning of the word with regard to the rest of the sentence. Combining these word representation vectors, results in a text representation matrix, which is then reduced to a single text representation vector, using the attention mechanism. The resulting text representation vector, or the bag-of-words, holds information of whether known words occur within the text, but not where they occur. This text representation vector is then classified.

Where the two classifier models differ, is the information that the word representation vectors they create contain:

\begin{itemize}
    \item For the baseline spaCy v2 TextCategorizer, the word representations created by Toc2Vec model are simpler, containing contextual information only found in the data provided (\url{https://spacy.io/api/tok2vec}).
    \item The spaCy v3 Robustly optimized Bidirectional Encoder Representations from Transformers (BERT) pretraining approach (RoBERTa) transformer TextCategorizer, has been pre-trained on large volumes of text data, and the word representations it creates holds rich contextual information, from data the model has been pre-trained with, appropriate to each context each word is found within (\url{https://spacy.io/api/transformer}).
\end{itemize}

Given the decision process the classifier goes through to make a classification prediction, an appropriate statistical text similarity analysis of the experiment data, would be to look at text representations comprised of the representations of the words that make up the text. While the vector representations created by the language models represent contextual information, Term-Frequency Inverse-Document-Frequency (TF-IDF) analysis vectors represent the importance of each word in a corpus, based on its frequency of appearance, in our case a data source and therapeutic domain specific dataset of posts.

% . . . . . . . . . . . . . . . . . . . . . . . . . . . . . . . . . . . . . . . . 
\subsubsection*{Term-Frequency Inverse-Document-Frequency and Cosine Similarity Analysis}

TF-IDF is a numerical statistic that captures how important or informative each word (or term) is within a dataset (\url{https://en.wikipedia.org/wiki/Tf–idf}):

\begin{equation}
    \begin{array}{ll}
         & TF\text{-}IDF(t) = TF(t) \cdot IDF(t) \vspace{0.25cm}\\ 
         & TF = \frac {\textrm{number of times the term appears in the document} }{ \textrm{total number of terms in the document}} \vspace{0.25cm}\\
         & IDF = log (\frac {\textrm{number of the documents in the corpus}} {\textrm{number of documents in the corpus contain the term}})
    \end{array}
\end{equation}

We use TF-IDF analysis to create vector representations of each data source and therapeutic domain specific dataset. Each vector representation of a dataset consists of the vector representations of all words or corpus of that dataset. Each word representation is the result of the TF-IDF statistic, assigning a value of significance to each word, based on the frequency of its occurrence within the dataset and across the datasets. Similarly to the classifier models' bag-of-words approach, the TF-IDF vector representations contain information on whether a word appears in the dataset, not where it appears.

We then compare the resulting TF-IDF vector representations of each dataset to identify the extent of how similar they are, when it comes to their corpus and each word's importance. To determine similarities of each possible pair of datasets in terms of their vocabularies, we perform a pairwise cosine similarity (cos) analysis:

\begin{align}
    cos(A, B) = \frac{A \cdot B}{||A|| * ||B||} = \frac{{\displaystyle\sum_{i=1}^{D}} a_i b_i}{\sqrt{{\displaystyle\sum_{i=1}^{D}} a_{i}^{2}} \sqrt{{\displaystyle\sum_{i=1}^{D}} b_{i}^{2}}} 
\end{align}

This is done by computing the dot product of both vectors (A and B) and normalizing it by the norm of both vectors. The dot product is computed by the sum of all pairs of elements multiplied, and the norm is the square root of the sum of all squared elements (e.g.~see \cite{hovy2020text}, pp.~39-40).

\begin{figure}[h]
\centering
\includegraphics[scale=0.75]{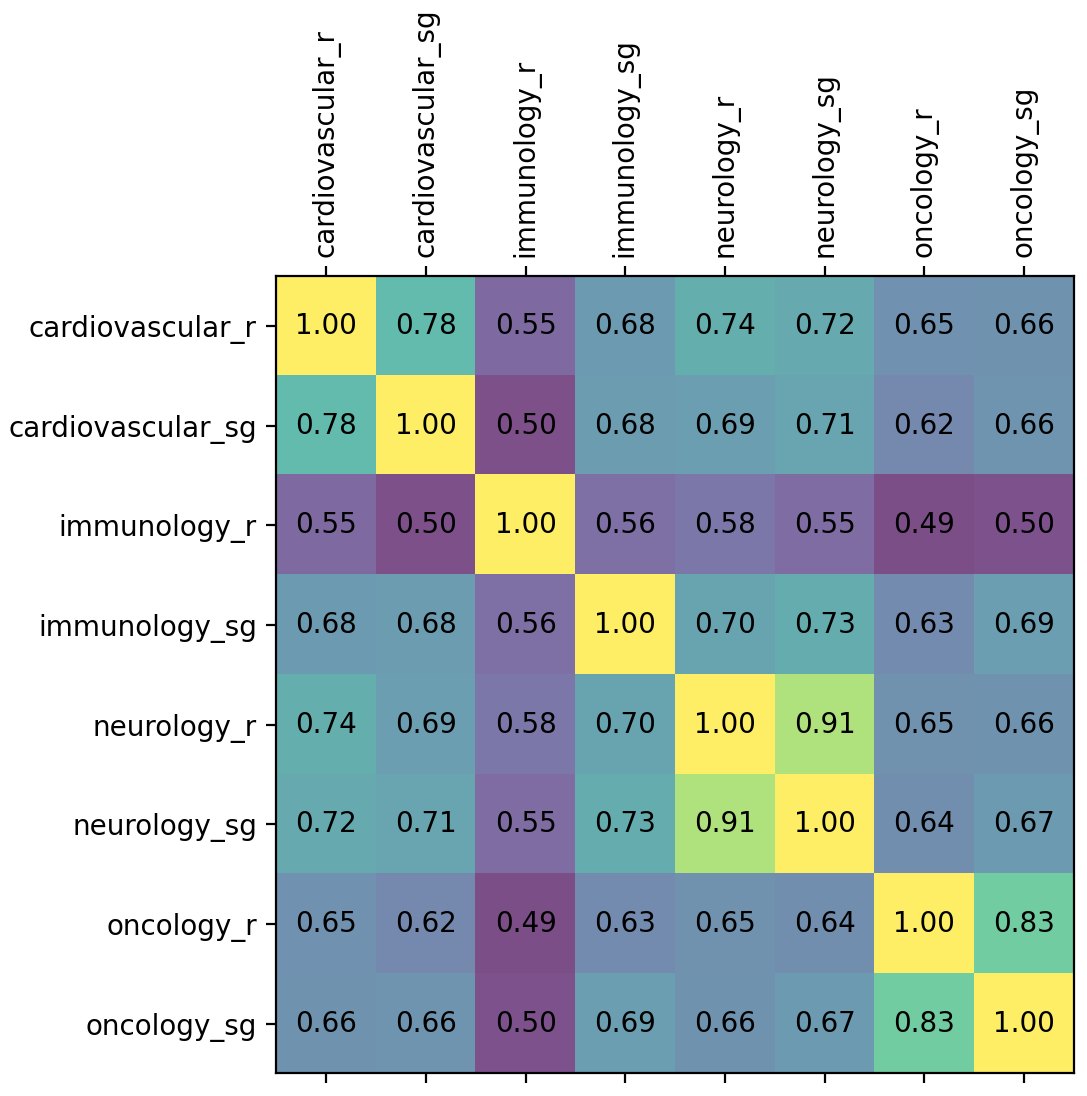}
\caption{Pairwise comparison matrix of cosine similarity values between all data source and therapeutic domain specific subsets of the data, after TF-IDF analysis.}
\label{fig:tf_idf}
\end{figure}

The result of this can be seen in Figure~\ref{fig:tf_idf}. A value of 1.0 and bright yellow colour signifies perfect similarity in vocabulary, while a lower value and darker blue colour represents lower similarity in the words occurring in both datasets. The analysis is performed on the stemmed corpus of each dataset, removing the stop words, which do not contribute much information, to better expose the similarities between the subsets of the data.

% . . . . . . . . . . . . . . . . . . . . . . . . . . . . . . . . . . . . . . . . 
\subsubsection*{Observations}

Referring to Figure~\ref{fig:tf_idf}, we consider a similarity of 0.45-0.60 to be low, 0.60-0.75 medium and anything above 0.75 considerable similarity between two datasets. Overall, most often, dataset pairs achieved medium similarity (18). Seven pairs showed low similarity, while three pairs showed significant similarity.

The patients' posts that share the most language in common, are ones within the same therapeutic domain across data sources. Between the two data sources (Reddit and SocialGist), cardiovascular, neurology and oncology patients use significantly similar language to express their experiences. Immunology patients on the other hand, use notably different language across the two data sources.

Two patient groups were observed to have consistently dissimilar language when compared to other groups. Reddit immunology patients' data (\texttt{immunology\_r}) achieved the lowest cosine similarity scores with other datasets in this study, even when compared to SocialGist immunology patients' language (\texttt{immunology\_sg}).

To better understand this observation, we extract for each dataset the top 20 words achieving the highest TF-IDF scores in our analysis, seen in descending order in Table~\ref{tab:tfidf_ranked}. Analysing these results, we confirm that Reddit immunology patients have the least similar vocabulary amongst all other patients, with 13 out of 20 top TF-IDF words being unique to the Reddit immunology dataset (compared to other dataset's top 20 TF-IDF score words), followed by SocialGist immunology with 10 out of 20 unique top TF-IDF words.

Immunology patients, as seen in Table~\ref{tab:tfidf_ranked}, are observed in SocialGist to use unique language to most commonly describe their medicines, like Humira, Ocrevus and Entyvio, while in Reddit they use uncommon words to describe their condition and body part, like eczema, skin and face. In comparison, all other therapeutic domain datasets have between 5 and 7 unique top 20 TF-IDF score words, with a lot of the common words between therapeutic domains being mentions of time like days, weeks, years, mentions of medical terminology like doctor, meds, treatment, and descriptions of their patient journeys like pain, help and effects.

% ////////////////////////////////////////////////////////////////////////////////
\subsection*{Experiments}

In this section, a series of experiments are described, aiming to optimally classify social media posts from Reddit and SocialGist by identifying patient voice. The experiments are comparisons between various subsets of the data as per their therapeutic domain or data source. Each experiment is performed using two different classifier model architectures, a strong baseline Convolutional Neural Network (CNN) model \cite{honnibal_2016} and a transformer model \cite{liu2019roberta}. We report performance of each classifier model trained on a data source and therapeutic domain specific dataset, or on a combination of these datasets.

% . . . . . . . . . . . . . . . . . . . . . . . . . . . . . . . . . . . . . . . . 
\subsubsection*{Experiment 1: Data source and therapeutic domain specific classifiers}

In the first experiment, we aim to capture nuances in language between patient groups, with data source and therapeutic domain specific classifiers. We train and evaluate a classifier for each therapeutic domain and data source, for each of the two model architectures. In the balance between having task specific classifiers to each patient group's language, and having general classifiers trained on broad volumes of data, this experiment leans on specificity.

The main comparison that can be made on the reported scores for this experiment (see Table~\ref{tab:tads_eval}) is between the two classifier architectures for each dataset as the same test set is used. The scores appearing in bold are the highest achieved in each metric (precision, recall, F1-score), for each test dataset, between the two classifier architectures. The highest achieved F1-scores between the two classifier architectures in each subset of the data range between 0.928 and 1.0, with the exception of the Reddit and SocialGist cardiovascular classifiers (\texttt{cardiovascular\_r}, \texttt{cardiovascular\_sg}), where the achieved F1-scores were lower at 0.865 and 0.760 respectively. The transformer classifier outperforms the CNN classifier in precision and F1-scores, while the two classifier architectures are tied in achieved recall scores.

% . . . . . . . . . . . . . . . . . . . . . . . . . . . . . . . . . . . . . . . . 
% \subsubsection*{Experiment 2: Therapeutic domain specific classifiers.}
\subsubsection*{Experiment 2: Combined therapeutic domain, data source and all data classifiers}

In the second experiment, we combine datasets belonging to the same therapeutic domain, e.g.~\texttt{cardiovascular\_r} and \texttt{cardiovascular\_sg} combined become \texttt{cardiovascular}, same data source, e.g. \texttt{cardiovascular\_r}, \texttt{oncology\_r}, \texttt{immunology\_r} and \texttt{neurology\_r} combined become \texttt{reddit\_coin}, and all datasets collected for this experiment, i.e.~\texttt{all}, to train and evaluate classifiers in each scenario, for each of the two model architectures. Informed by the TF-IDF analysis, the similarities in vocabulary between datasets, e.g.~datasets in the same therapeutic domain, could allow combining these datasets, training classifiers with more data to achieve higher evaluation scores. These experiments test having more general classifiers, trained on most, and all of the available data. As in Experiment 1, the main comparison that can be made on the reported scores for this experiment is between the two classifier architectures for each combined dataset.

The results of Experiment 2 are seen in Table~\ref{tab:combined_eval}, where the scores appearing in bold are the highest achieved in each metric (precision, recall, F1-score) for each test dataset between the two classifier architectures. The highest achieved F1-scores between the two classifier architectures in each subset of the data range between 0.911 and 0.948, with the former obtained by the \texttt{immunology} and \texttt{socialgist\_coin} classifiers. The low cosine similarity score between the two immunology datasets revealed in the TF-IDF analysis, partly explains the comparatively low F1-score achieved by the \texttt{immunology} classifier. The transformer classifier slightly outperforms the CNN classifier in recall and F1-scores, while the two classifier architectures are tied in precision scores.

% . . . . . . . . . . . . . . . . . . . . . . . . . . . . . . . . . . . . . . . . 
% \subsubsection*{Experiment 3: Data source specific classifiers.}
\subsubsection*{Experiment 3: All classifiers comparison on each therapeutic domain and data source specific dataset.}

The third and final experiment compares the performance of all classifiers containing each therapeutic domain and data source specific dataset in their train data, evaluated on each dataset's test set. For each therapeutic domain and data source specific dataset, this comparison shows whether the best performing classifier is the therapeutic domain data source specific one, the therapeutic domain, the data source or the all data one. An example would be the Reddit cardiovascular dataset, which is part of the \texttt{cardiovascular\_r}, \texttt{cardiovascular}, \texttt{reddit\_coin} and \texttt{all} classifiers' train data, so all of these classifiers are evaluated on the same Reddit cardiovascular test dataset.

The results of this comprehensive comparison can be seen in Table~\ref{tab:all_eval}, where the scores appearing in bold are the highest achieved in each metric (precision, recall, F1-score) for each therapeutic domain data source specific test dataset, between the two classifier architectures and the therapeutic domain data source specific, therapeutic domain specific, data source specific or all data classifier. The highest achieved F1-scores between the four test datasets range between 0.977 and 1.0, with the exception of the Reddit and SocialGist cardiovascular test datasets, where the highest achieved F1-scores were lower at 0.865 and 0.863 respectively. The transformer classifier outperforms the CNN classifier in all precision, recall and F1-score comparisons, with the exception of Reddit neurology recall, where the best CNN classifier outperforms the transformer classifier by a negligible margin.

% ////////////////////////////////////////////////////////////////////////////////

\section*{Methods}

% . . . . . . . . . . . . . . . . . . . . . . . . . . . . . . . . . . . . . . . . 
\subsection*{Text classifier models}

We use spaCy's TextCategorizer module for the modelling needs of this work. It supports multiple language model architectures, which use a bag-of-words approach to classify text. We train the models on our training data, seen in Table~\ref{tab:experiment_data_volumes} and \ref{tab:experiment_combined_data_volumes}, for our specific task of identifying patient voice amongst online posts.

\textit{Baseline spaCy CNN TextCategorizer}: As a baseline for this experiment, we use spaCy's small CNN TextCategorizer \texttt{en\_core\_web\_sm}, for multi-label text classification. As per spaCy's documentation, the ensemble small CNN model consists of a neural network and a CNN model, with mean pooling and attention. It is a strong baseline, balancing efficiency and accuracy (\url{https://spacy.io/models/en#en_core_web_sm}). We refer to it throughout the experiments of this paper as the "CNN" classifier.

\textit{Transformer RoBERTa base TextCategorizer}: SpaCy's TextCategorizer offers a more accurate solution, at the expense of additional computational requirements, the transformer language model \texttt{en\_core\_web\_trf}. It consists of the RoBERTa base language model \cite{liu2019roberta}, that is pre-trained on a large corpus of text, and is available for fine-tuning, in this case to perform the downstream task of text classification (\url{https://spacy.io/models/en#en_core_web_trf}). We refer to it throughout the experiments of this paper as the "transformer" classifier.

% . . . . . . . . . . . . . . . . . . . . . . . . . . . . . . . . . . . . . . . . 
\subsection*{Evaluation Metrics}

For document classification as well as inter-annotator agreement performance, we report the standard metrics of precision, recall and F1-score for each label type, weighted and macro averaged F1-score across all label types. In IAA, we also report Cohen's kappa, as a more robust metric of agreement. The classifier models are evaluated on the test datasets, with their details seen in Tables~\ref{tab:experiment_data_volumes} and~\ref{tab:experiment_combined_data_volumes}.

% ≡≡≡≡≡≡≡≡≡≡≡≡≡≡≡≡≡≡≡≡≡≡≡≡≡≡≡≡≡≡≡≡≡≡≡≡≡≡≡≡≡≡≡≡≡≡≡≡≡≡≡≡≡≡≡≡≡≡≡≡≡≡≡≡≡≡≡≡≡≡≡≡≡≡≡≡≡≡≡≡

% \section*{Methods}
\section*{Discussion}

% Topical subheadings are allowed. Authors must ensure that their Methods section includes adequate experimental and characterization data necessary for others in the field to reproduce their work.

Our initial linguistic analysis, surfaced differences in patient language across data sources and therapeutic domains, beyond the expected domain specific medical terminology. Patients on message boards (SocialGist data) appeared to write longer posts, using a richer vocabulary to describe their experiences, compared to patient posts coming from social media (Reddit).

Patients posting on cardiovascular related conditions mention specific dosages, side effects and lifestyle factors. This could be because these individuals often undergo one or at least very few life events that have resulted in their cardiovascular diagnosis e.g. heart attack or stroke. Often cardiovascular patients remain unaware of an underlying health condition until it is too late. Their accounts focus on life before this event – the traits they possessed at that period of time (e.g. weight characteristics) compared to their post diagnosis state. In the rarer occasion that patients are made aware of their symptoms ahead of crisis point, they still cannot feel these symptoms themselves to the same extent an immunology patient with itchy skin could. High blood pressure for instance would be almost impossible to detect for oneself without aid of a wearable device. When medicines are discussed in this cardiovascular context, the patient conversation often focusses on the side effects of these medicines, that can be detected by a patient, rather than the disease related symptoms.

Oncology patients are also reliant to some extent on conversations they have with doctors and specialists to inform them of their diagnosis and disease state. Our linguistic analysis determined that conversations around oncologists were prevalent in these datasets. Despite this similarity to cardiovascular patient language, there are still significant differences between how these two groups of patients speak. In this experiment, oncology patient accounts contained the most concise and highly specific language. Individuals often talk about specific genes and attributes, as well as clinical trials and gene tests that relate to these.

Conversely to cardiovascular and oncology patients, those suffering from immunological diseases are often the best placed to give elaborate details on their conditions. The symptoms experienced are highly noticeable and can impact day to day quality of life in a considerable way. These patient posts regularly go into a large amount of detail about the physical symptoms of their condition – talking about body parts affected and exact sensations experienced. These individuals can qualitatively describe the impact their condition has had on them both mentally and physically.

Neurology patients can also be seen to subjectively describe their quality of life at given periods in time relating to different medicines and stages in their journey. The majority of the discussion in these posts is around feelings and emotions.

Performing a statistical text similarity analysis, TF-IDF confirmed these differences in patient language across data sources and therapeutic domains. We observed that the corpus of patients in each domain, after removing stop words that do not convey significant meaning, showed similarities only within the same therapeutic domain, across the two data sources, except for immunology. The language between Reddit and SocialGist immunology patients was quite dissimilar. In fact, the Reddit immunology dataset was quite dissimilar to all other datasets in this experiment. This is reflected in the evaluation results in Table~\ref{tab:combined_eval}. Of the classifiers trained on the combined therapeutic domain data from both data sources, the immunology classifier achieves the lowest F1-score of the four at 0.911. The statistical text similarity analysis overall provided us with useful information on which datasets would be appropriate to combine to prior to the experiments, while it also provided partial explanations on the classifiers' performance when trained and evaluated on combined datasets.

The classifier modelling experiments reflect our observations made in the linguistic and statistical text analysis. The TF-IDF similarity analysis showed similarity between patients' language within the same therapeutic domain across both data sources. Observing the achieved F1-scores in Table~\ref{tab:all_eval}, the classifiers trained on combined datasets, outperform the classifiers trained on therapeutic domain and data source specific datasets, with the exception of the Reddit cardiovascular and Reddit neurology classifiers which slightly outperformed classifiers trained on combined datasets. These results confirm that if TF-IDF similarity analysis identifies similarities between patients' language between datasets, it is worth combining the datasets into a larger one and training a single classifier that outperforms classifiers trained on each individual dataset.

Our analysis of the results of Experiment 3 as seen in Table~\ref{tab:all_eval}, reveals two things. The transformer model architecture consistently outperforms the CNN model architecture, in all three metrics of precision, recall and F1-score. The same comparison of model architectures can be made for the results shown in Tables~\ref{tab:tads_eval} and~\ref{tab:combined_eval}. There we made the same observation that the transformer classifiers also outperform the CNN classifiers in all three evaluation metrics, although not consistently. This can to a large extent be attributed to the transformer model's pre-training. Additionally, Table~\ref{tab:all_eval} shows that transformer classifiers trained on more data, achieved higher F1-scores than classifiers trained on fewer data. Specifically, the \texttt{all} classifier has achieved the highest F1-scores in all of the comparisons it has been part of. An exception to this is the Reddit cardiovascular and Reddit neurology test datasets, where the \texttt{all} classifier was outperformed by the therapeutic domain and data source specific classifiers, by a minor margin. 

On the other hand, the CNN classifier model, as seen in the comparisons in Table~\ref{tab:all_eval}, achieved highest F1-scores most often for the data source specific classifiers, followed by the therapy area specific classifiers, with the highly specific or general all data classifier coming last. Overall, we observed that larger and more complex transformer language models, trained on more data, are the best performing classifiers in classifying patients posts collected from social media and message boards.

\section*{Conclusions}

In this paper, we report on a series of experiments to determine optimal identification of patient voice from posts collected from social media and message boards. This is done across four therapeutic domains, cardiovascular, oncology, immunology and neurology, each one with its own linguistic features. We performed a linguistic and statistical text similarity analysis, identifying the distinct ways patients use language to express their experiences across therapeutic domains and data sources. Using TF-IDF similarity analysis, we identify which datasets are more similar in vocabulary than others, informing us for which datasets it is beneficial to combine them into one for model training, and helping interpret the performance scores achieved by the classifiers. With the exception of immunology, patients in each therapeutic domain use similar language across the two data sources to talk about their experiences.

We compute IAA to determine the difficulty of patient voice classification and to get insight into the upper bound of automatic classification performance for this task. The achieved IAA scores show significant agreement, with a weighted averaged F1-score of 0.952 and a macro averaged F1-score of 0.883. The mean Cohen's Kappa score of agreement between all pairs of annotators is 0.773, suggesting substantial agreement. The data is then split into train and validation sets, using an 80\% train and 20\% validation split, with additional test data collected equal to the 20\% validation set, used for model evaluation (Tables~\ref{tab:experiment_data_volumes} and~\ref{tab:experiment_combined_data_volumes}). We train a classifier model on each subset of the data, or a combination of datasets. Two classifier architectures are compared, a CNN and a transformer classifier model.

The results show that the linguistic analysis was able to identify similarities between datasets, specifically with respect to the language used by patients across the two data sources in each therapeutic domain. These similarities were then leveraged by combining the datasets, in order to train classifier models with more data, yielding higher performance than their classifier counterparts trained on fewer data. To obtain optimal classifier performance for the task of identifying patient voice in online social media and message board posts, this work shows that:

\begin{itemize}
    \item Linguistic and statistical text similarity analysis can be used to identify similar datasets that can be combined into larger datasets.
    \item Text classifier models trained on more data outperform classifier models trained on less data.
    \item Pre-trained transformer models outperform CNN models as bag-of-words text classifiers for the task of patient voice classification.
\end{itemize}

% GL suggested conclusion
We confirm our initial hypothesis, that understanding the linguistic characteristics of each dataset, across different data sources and therapeutic domains, achieves more accurate patient voice classification. By combining datasets with similar language and creating larger datasets for training, the models' ability to perform is improved.

\section*{Data availability}

We provide the list of subreddits and search terms, which we used to collect the data for this research and development project, in the supplementary material. The annotation labels and examples are also described in this paper. The third-party tools (classifiers and annotation tool) used for this work are freely available and details on the classifier set-up and model parameters are provided in this paper. For more information about this project and the data please contact Elizabeth A.L. Fairley.

\bibliography{bibliography}

% \noindent LaTeX formats citations and references automatically using the bibliography records in your .bib file, which you can edit via the project menu. Use the cite command for an inline citation, e.g.  \cite{Hao:gidmaps:2014}.

% For data citations of datasets uploaded to e.g. \emph{figshare}, please use the \verb|howpublished| option in the bib entry to specify the platform and the link, as in the \verb|Hao:gidmaps:2014| example in the sample bibliography file.

\section*{Acknowledgements}

% Acknowledgements should be brief, and should not include thanks to anonymous referees and editors, or effusive comments. Grant or contribution numbers may be acknowledged.

We would like to express our gratitude to the invaluable Talking Medicines Limited annotators for their hard work in creating the data needed for model training and validation. We would also like to thank Ellen Halliday, for her consultation in all compliance matters concerning this paper's experiments, as well as the Talking Medicines Founders Jo-Anne Halliday, Scott F. Crae and Elizabeth A.L. Fairley for their support of this project.

\section*{Author contributions}

% Must include all authors, identified by initials, for example:
% A.A. conceived the experiment(s),  A.A. and B.A. conducted the experiment(s), C.A. and D.A. analysed the results.  All authors reviewed the manuscript. 

G.L. and R.E.O. co-wrote this paper. R.E.O. managed the data annotation used for training, validation, evaluation and inter-annotator agreement calculations, and performed the linguistic analysis on the data of this experiment. G.L. performed the statistical text similarity analysis, IAA calculations and modelling experiments. V.P. sourced the experiment data, processed it for annotation and set up the tools to annotate it. G.L.B supported on engineering infrastructure to run the experiments. B.A. advised on aspects of the work involved in this project, and edited the paper. E.A.L.F. advised on the overall direction of the project and edited the paper.

\section*{Funding}

This work was funded by Talking Medicines Limited.

\section*{Additional information}

\subsection*{Competing interests}

The authors declare no competing interests.

% To include, in this order: \textbf{Accession codes} (where applicable); \textbf{Competing interests} (mandatory statement). 

% The corresponding author is responsible for submitting a \href{http://www.nature.com/srep/policies/index.html#competing}{competing interests statement} on behalf of all authors of the paper. This statement must be included in the submitted article file.

% \begin{figure}[ht]
% \centering
% \includegraphics[width=\linewidth]{stream}
% \caption{Legend (350 words max). Example legend text.}
% \label{fig:stream}
% \end{figure}

% \begin{table}[ht]
% \centering
% \begin{tabular}{|l|l|l|}
% \hline
% Condition & n & p \\
% \hline
% A & 5 & 0.1 \\
% \hline
% B & 10 & 0.01 \\
% \hline
% \end{tabular}
% \caption{\label{tab:example}Legend (350 words max). Example legend text.}
% \end{table}

% Figures and tables can be referenced in LaTeX using the ref command, e.g. Figure \ref{fig:stream} and Table \ref{tab:example}.

\section*{Tables}

\renewcommand{\arraystretch}{1.25}

\begin{table}[htpb]
\centering
\begin{tabular}{llllllll}
\hline
\multicolumn{2}{c}{Cardiovascular} & \multicolumn{2}{c}{Oncology}         & \multicolumn{2}{c}{Immunology}  & \multicolumn{2}{c}{Neurology}    \\
Reddit          & SocialGist       & Reddit             & SocialGist         & Reddit             & SocialGist         & Reddit          & SocialGist       \\ \hline
stroke          & \textbf{blood}   & \textbf{arimidex}  & \textbf{chemo}     & eczema             & \textbf{years}     & \textbf{feel}   & \textbf{abilify} \\
\textbf{heart}  & eliquis          & week               & \textbf{herceptin} & skin               & humira             & \textbf{taking} & \textbf{zyprexa} \\
cholesterol     & \textbf{heart}   & test               & \textbf{treatment} & \textbf{psoriasis} & entyvio            & \textbf{really} & \textbf{time}    \\
\textbf{know}   & \textbf{day}     & \textbf{cancer}    & \textbf{years}     & face               & \textbf{time}            & zoloft          & \textbf{taking}  \\
\textbf{blood}  & \textbf{years}   & \textbf{chemo}   & \textbf{cancer}   & \textbf{really} & tysabri & \textbf{abilify} & \textbf{meds} \\
left            & xarelto          & \textbf{weeks}     & \textbf{good}      & \textbf{know}      & \textbf{good}            & \textbf{time}   & \textbf{years}   \\
right           & \textbf{effects} & \textbf{herceptin} & \textbf{time}      & use                & remicade      & vyvanse         & topamax          \\
\textbf{high}   & entresto         & cycle              & \textbf{effects}   & cream              & \textbf{weeks}           & \textbf{day}    & \textbf{day}     \\
\textbf{years}  & \textbf{time}    & \textbf{treatment} & hope               & \textbf{time}      & \textbf{know}     & latuda          & \textbf{effects} \\
\textbf{time}   & \textbf{doctor}  & \textbf{time}    & \textbf{know}     & \textbf{years}  & stelara & \textbf{zyprexa} & \textbf{feel} \\
\textbf{feel}   & \textbf{said}    & perjeta            & \textbf{avastin}   & flare              & \textbf{effects}            & \textbf{help}   & \textbf{good}    \\
normal          & \textbf{know}    & \textbf{know}      & faslodex           & \textbf{feel}      & \textbf{months}   & \textbf{meds}   & \textbf{know}    \\
\textbf{doctor} & \textbf{taking}  & \textbf{avastin}   & \textbf{feel}      & bad                & \textbf{started}    & \textbf{years}  & mg               \\
pressure        & coumadin         & days               & \textbf{weeks}     & \textbf{help}      & cosentyx   & champix         & \textbf{work}    \\
\textbf{help}   & warfarin         & \textbf{months}    & phesgo             & body               & enbrel           & anxiety         & \textbf{weight}  \\
\textbf{ago}    & \textbf{ago}     & e2                 & oncologist         & dry                & \textbf{psoriasis}             & \textbf{know}   & symptoms         \\
\textbf{pain}   & \textbf{think}   & \textbf{good}      & \textbf{said}      & itchy              & infusion & \textbf{work}   & sleep            \\
hospital        & \textbf{pain}    & \textbf{taking}    & \textbf{months}    & scalp              & \textbf{pain}           & efexor          & tried            \\
\textbf{said}   & \textbf{high}    & \textbf{started} & \textbf{arimidex} & worse           & drug & \textbf{effects} & risperdal     \\
\textbf{think}  & \textbf{meds}    & \textbf{effects}   & scan               & hands              & biologic               & \textbf{weight} & migraines        \\ \hline
\end{tabular}
\caption{Top 20 words per dataset, ranked by their TF-IDF score in descending order, with bold denoting words common in more than one dataset's top 20 words list.}
\label{tab:tfidf_ranked}
\end{table}

\begin{table}[htpb]
\centering
\begin{tabular}{llllllll}
\hline
\multicolumn{2}{l}{\multirow{2}{*}{Classifier}} & \multicolumn{3}{c}{CNN}                          & \multicolumn{3}{c}{Transformer}                  \\
% \multicolumn{2}{l}{}                            & P              & R              & F1             & P              & R              & F1             \\ \hline
\multicolumn{2}{l}{}                            & precision       & recall         & F1-score       & precision       & recall       & F1-score      \\ \hline
\multirow{2}{*}{Cardiovascular} & Reddit & 0.742 & \textbf{0.915} & 0.819          & \textbf{0.839} & 0.891 & \textbf{0.865} \\
                               & SocialGist     & \textbf{0.635} & 0.946          & \textbf{0.760} & 0.513          & \textbf{0.983} & 0.674          \\ \cline{2-8} 
\multirow{2}{*}{Oncology}      & Reddit         & 0.945          & \textbf{0.986} & \textbf{0.980} & \textbf{1.0}   & 0.808          & 0.894          \\
                               & SocialGist     & 0.947          & \textbf{0.971} & 0.959          & \textbf{0.970} & 0.953          & \textbf{0.961} \\ \cline{2-8} 
\multirow{2}{*}{Immunology}     & Reddit & 0.947 & \textbf{0.910} & \textbf{0.928} & \textbf{0.972} & 0.881 & 0.924          \\
                               & SocialGist     & 0.959          & 0.977          & 0.968          & \textbf{0.977} & \textbf{0.981} & \textbf{0.979} \\ \cline{2-8} 
\multirow{2}{*}{Neurology}     & Reddit         & 0.995          & 0.973          & 0.984          & \textbf{1.0}   & \textbf{1.0}   & \textbf{1.0}   \\
                               & SocialGist     & 0.693          & 0.972          & 0.809          & \textbf{0.894} & \textbf{1.0}   & \textbf{0.944} \\ \hline
\end{tabular}
\caption{Data source and therapeutic domain specific classifiers precision, recall and F1-score evaluation scores, each classifier evaluated on their own test datasets.}
\label{tab:tads_eval}
\end{table}

\begin{table}[htpb]
\centering
\begin{tabular}{lllllll}
\hline
\multirow{2}{*}{Classifier} & \multicolumn{3}{c}{CNN}                          & \multicolumn{3}{c}{Transformer}                  \\
                            % & P              & R              & F1             & P              & R              & F1             \\ \hline
                            & precision      & recall         & F1-score       & precision      & recall         & F1-score       \\ \hline
Cardiovascular              & 0.896          & \textbf{0.962} & 0.928          & \textbf{0.921} & 0.960          & \textbf{0.940} \\
Oncology                    & \textbf{0.936} & 0.961          & \textbf{0.948} & 0.857          & \textbf{0.991} & 0.919          \\
Immunology                  & \textbf{0.865} & 0.934          & 0.898          & 0.845          & \textbf{0.989} & \textbf{0.911} \\
Neurology                   & 0.920          & \textbf{0.947} & \textbf{0.933} & \textbf{0.959} & 0.878          & 0.917          \\ \cline{2-7}
Reddit COIN                 & \textbf{0.892} & 0.919          & 0.905          & \textbf{0.892} & \textbf{0.971} & \textbf{0.930} \\
SocialGist COIN             & \textbf{0.940} & 0.884          & \textbf{0.911} & 0.676          & \textbf{1.0}   & 0.806          \\ \cline{2-7}
All                         & 0.866          & \textbf{0.968} & 0.915          & \textbf{0.942} & 0.936          & \textbf{0.939} \\ \hline
\end{tabular}
\caption{Data source specific, therapeutic domain specific and all data classifiers precision, recall and F1-score evaluation scores, each classifier evaluated on their own test datasets.}
\label{tab:combined_eval}
\end{table}

\begin{table}[htpb]
\centering
\begin{tabular}{lllllllll}
\hline
% \multicolumn{2}{l}{\multirow{2}{*}{Test Dataset}}             & \multirow{2}{*}{Classifier} & \multicolumn{3}{c}{CNN} & \multicolumn{3}{c}{Transformer}                \\
\multicolumn{2}{l}{Test}             & \multirow{2}{*}{Classifier} & \multicolumn{3}{c}{CNN} & \multicolumn{3}{c}{Transformer}                \\
% & \multirow{2}{*}{Test Dataset}             & \multirow{2}{*}{Classifier} & \multicolumn{3}{c}{CNN} & \multicolumn{3}{c}{Transformer}                \\
% \multicolumn{2}{l}{}           &                           & P     & R            & F1    & P              & R              & F1             \\ \hline
\multicolumn{2}{l}{Dataset}           &                           & precision     & recall            & F1-score    & precision              & recall              & F1-score             \\ \hline
\multirow{8}{*}{\rotatebox[origin=c]{90}{Cardiovascular}} & \multirow{4}{*}{\rotatebox[origin=c]{90}{Reddit}}     & Cardiovascular Reddit       & 0.742  & 0.915  & 0.819 & \textbf{0.839} & 0.891        & \textbf{0.865} \\
 &                             & Cardiovascular            & 0.738 & 0.938        & 0.826 & 0.726          & \textbf{0.977} & 0.833          \\
 &                             & Reddit COIN               & 0.751 & 0.922        & 0.828 & 0.740          & 0.969          & 0.839          \\
 &                             & All                       & 0.719 & 0.961        & 0.823 & 0.790          & 0.950          & 0.863          \\ \cline{2-9} 
 & \multirow{4}{*}{\rotatebox[origin=c]{90}{SocialGist}} & Cardiovascular SocialGist & 0.635 & 0.946        & 0.760 & 0.513          & 0.983          & 0.674          \\
 &                             & Cardiovascular            & 0.738 & 0.938        & 0.826 & 0.726          & 0.977          & 0.833          \\
 &                             & SocialGist COIN           & 0.679 & 0.969        & 0.799 & 0.516          & \textbf{1.0}   & 0.681          \\
 &                             & All                       & 0.719 & 0.961        & 0.823 & \textbf{0.790} & 0.950          & \textbf{0.863} \\ \hline
\multirow{8}{*}{\rotatebox[origin=c]{90}{Oncology}}       & \multirow{4}{*}{\rotatebox[origin=c]{90}{Reddit}}     & Oncology Reddit             & 0.945  & 0.986  & 0.980 & \textbf{1.0}   & 0.808        & 0.894          \\
 &                             & Oncology                  & 0.961 & 0.989        & 0.975 & 0.962          & 0.996          & 0.979          \\
 &                             & Reddit COIN               & 0.967 & 0.964        & 0.966 & 0.982          & \textbf{1.0}   & 0.991          \\
 &                             & All                       & 0.955 & 0.996        & 0.975 & \textbf{1.0}   & 0.986          & \textbf{0.993} \\ \cline{2-9} 
 & \multirow{4}{*}{\rotatebox[origin=c]{90}{SocialGist}} & Oncology SocialGist       & 0.947 & 0.971        & 0.959 & 0.970          & 0.953          & 0.961          \\
 &                             & Oncology                  & 0.947 & 0.975        & 0.961 & 0.932          & 0.996          & 0.963          \\
 &                             & SocialGist COIN           & 0.948 & 0.993        & 0.970 & 0.923          & \textbf{1.0}   & 0.960          \\
 &                             & All                       & 0.938 & 0.996        & 0.967 & \textbf{0.971} & 0.989          & \textbf{0.980} \\ \hline
\multirow{8}{*}{\rotatebox[origin=c]{90}{Immunology}}     & \multirow{4}{*}{\rotatebox[origin=c]{90}{Reddit}}     & Immunology Reddit           & 0.947  & 0.910  & 0.928 & \textbf{0.972} & 0.881        & 0.924          \\
 &                             & Immunology                & 0.963 & 0.939        & 0.951 & 0.949          & \textbf{1.0}   & 0.974          \\
 &                             & Reddit COIN               & 0.957 & 0.975        & 0.966 & 0.965          & 0.986          & 0.975          \\
 &                             & All                       & 0.963 & 0.931        & 0.947 & 0.968          & 0.986          & \textbf{0.977} \\ \cline{2-9} 
 & \multirow{4}{*}{\rotatebox[origin=c]{90}{SocialGist}} & Immunology SocialGist     & 0.959 & 0.977        & 0.968 & 0.977          & 0.981          & 0.979          \\
 &                             & Immunology                & 0.943 & 0.950        & 0.947 & 0.956          & 0.996          & 0.976          \\
 &                             & SocialGist COIN           & 0.952 & 0.989        & 0.970 & 0.900          & \textbf{1.0}   & 0.947          \\
 &                             & All                       & 0.959 & 0.985        & 0.972 & \textbf{0.981} & 0.981          & \textbf{0.981} \\ \hline
\multirow{8}{*}{\rotatebox[origin=c]{90}{Neurology}}      & \multirow{4}{*}{\rotatebox[origin=c]{90}{Reddit}}     & Neurology Reddit            & 0.995  & 0.973  & 0.984 & \textbf{1.0}   & \textbf{1.0} & \textbf{1.0}   \\
 &                             & Neurology                 & 0.995 & 0.973        & 0.984 & \textbf{1.0}   & 0.979          & 0.989          \\
 &                             & Reddit COIN               & 0.979 & 0.995        & 0.987 & 0.995          & 0.995          & 0.995          \\
 &                             & All                       & 0.954 & \textbf{1.0} & 0.977 & 0.995          & 0.995          & 0.995          \\ \cline{2-9} 
                                & \multirow{4}{*}{\rotatebox[origin=c]{90}{SocialGist}} & Neurology SocialGist        & 0.693  & 0.972  & 0.809 & 0.894          & \textbf{1.0} & 0.944          \\
 &                             & Neurology                 & 0.814 & 0.944        & 0.875 & \textbf{1.0}   & 0.986          & \textbf{0.993} \\
 &                             & SocialGist COIN           & 0.781 & 0.993        & 0.875 & 0.571          & \textbf{1.0}   & 0.727          \\
 &                             & All                       & 0.753 & 0.993        & 0.856 & 0.986          & \textbf{1.0}   & \textbf{0.993} \\ \hline
\end{tabular}
\caption{All experiment classifiers, evaluated on therapeutic domain and data source specific test datasets. For each test dataset, the precision, recall and F1-score are compared between the all data classifier, the data source specific classifier, the therapeutic domain classifier, and the therapeutic domain and data source specific classifier.}
\label{tab:all_eval}
\end{table}

\end{document}